%% file: acl_latex.tex
\pdfoutput=1

\documentclass[11pt]{article}

\usepackage{acl}

\usepackage{times}
\usepackage{latexsym}
\usepackage{tikz}
\usepackage{tikz-qtree}
\usetikzlibrary{bayesnet}
\usetikzlibrary{trees,shapes,arrows}
\usepackage[switch]{lineno}
\usepackage[edges]{forest}
\usepackage[T1]{fontenc}

\usepackage[utf8]{inputenc}
\usepackage{multirow}
\usepackage{booktabs}

\definecolor{hidden-red}{RGB}{205, 44, 36}
\definecolor{hidden-blue}{RGB}{228,245,252}

\definecolor{hidden-orange}{RGB}{243,202,120}
\definecolor{hidden-green}{RGB}{34,139,34}
\definecolor{hidden-pink}{RGB}{255,245,247}
\definecolor{hidden-black}{RGB}{20,68,106}

\usepackage{microtype}
\usepackage{graphicx}
\usepackage{rotating}
\usepackage{pdflscape}
\usepackage{subcaption}
\usepackage{rotfloat}
\usepackage{minitoc}

\usepackage{hyperref}
\usepackage{url} 
\title{Video-Language Understanding: A Survey from Model Architecture, Model Training, and Data Perspectives}

\author{~~Thong Nguyen$^{1}$,~~Yi Bin$^{1}$\thanks{~~Corresponding author}~~,~~Junbin Xiao$^{1}$,~~\textbf{Leigang Qu}$^{1}$,~~\textbf{Yicong Li}$^{1}$,\\~~\textbf{Jay Zhangjie Wu}$^{1}$, ~~\textbf{Cong-Duy Nguyen}$^{2}$,~~\textbf{See--Kiong Ng}$^{1}$,~~\textbf{Luu Anh Tuan}$^{2}$$^{*}$ \\
  $^1$National University of Singapore, Singapore \\
  $^2$Nanyang Technological University, Singapore \\
    \texttt{\small thong.nguyen@u.nus.edu, anhtuan.luu@ntu.edu.sg} \\}

\begin{document}
\maketitle
\begin{abstract}

Humans use multiple senses to comprehend the environment. Vision and language are two of the most vital senses since they allow us to easily communicate our thoughts and perceive the world around us. There has been a lot of interest in creating video-language understanding systems with human-like senses since a video-language pair can mimic both our linguistic medium and visual environment with temporal dynamics. In this survey, we review the key tasks of these systems and highlight the associated challenges. Based on the challenges, we summarize their methods from model architecture, model training, and data perspectives. We also conduct performance comparison among the methods, and discuss promising directions for future research. 
\end{abstract}

\input{folders/01_introduction}
\input{folders/02_tasks}
\input{folders/03_challenges}
\input{folders/04_model_architecture}
\input{folders/05_model_training}
\input{folders/06_data}
\input{folders/07_future_directions}
\input{folders/08_conclusion}

\input{folders/09_limitations}
\input{folders/10_acknowledgement}

\bibliography{anthology,custom}
\bibliographystyle{acl_natbib}

\clearpage
\appendix
\input{folders/appendix}

\end{document}

%% file: folders/01_introduction.tex
\section{Introduction}
Vision and language constitute fundamental components of our perception: vision allows us to perceive the physical world, while language enables us to describe and converse about it. However, the world is not merely a static image but exhibits dynamics in which objects move and interact across time. With the temporal dimension, videos are able to capture such temporal dynamics that characterize the physical world. Consequently, in pursuit of endowing artificial intelligence with human-like perceptual abilities, researchers have been developing Video-Language Understanding models that are capable of interpreting the spatio-temporal dynamics of videos and the semantics of language, dating back to the 1970s \citep{lazarus1973multimodal, mcgurk1976hearing}. These models are distinctive from image-language understanding models, since they exhibit an additional ability to interpret the temporal dynamics \citep{li2020hero}.

They have demonstrated impressive performance in various video-language understanding tasks. These tasks evaluate video-language models from coarse-grained to fine-grained understanding capacity. For example, for coarse-grained understanding, text-video retrieval task assesses the model’s ability to holistically associate a language query with a whole video \citep{han2023bic}. For more fine-grained understanding capacity, a video captioning model is required to understand the overall and detailed video content, then describe the content in concise language \citep{abdar2023review}. Fine-grained understanding in video questioning answering remains a difficult task, where a model needs to recognize minute visual objects or actions, and infers their semantic, spatial, temporal, and causal relationships \citep{xiao2021next}.  

In order to effectively perform such video-language understanding tasks, there are three challenges that video-language understanding works have to explore. The first challenge lies in devising an appropriate neural architecture to model the interaction between video and language modalities. The second challenge is to design an effective strategy to train video-language understanding models in order to effectively adapt to multiple target tasks and domains. The third challenge is preparing high-quality video-language data that fuel the training of these models.
\input{folders/taxonomy}

Although a handful of recent works have tried to review video-language understanding, they mostly focus on one challenge, for example, Transformer-based \citep{ruan2022survey} and LLM-augmented architecture \citep{tang2023video} (the 1st challenge), self-supervised learning \citep{schiappa2023self} and pre-training  \citep{cheng2023vindlu} (the 2nd challenge), and data augmentation \citep{zhou2024survey} (the 3rd challenge). Moreover, others also focus merely on one video-language understanding task, \textit{e.g.} video question answering \citep{zhong2022video}, text-video retrieval \citep{zhu2023deep}, and video captioning \citep{abdar2023review}. 
Such a narrow focus contradicts the growing consensus advocating for the development of artificial general intelligence capable of versatile adaptation to a range of tasks and domains. Consider a human interaction scenario where an individual iteratively poses questions about a video, searches for a pertinent moment, and requests a summary. Such use case necessitates a broad capability to comprehend video and language content, without being bounded by a certain task. In addition, the development of a video-language understanding system often involves a multi-step process encompassing designing a model architecture, formulating a training method, and preparing data, rather than being a singular-step endeavor. Hence, this paper aims to present a full-fledged and meaningful survey to connect the aspects of video-language understanding. Our contributions are as follows:
\begin{itemize}
    \item We summarize the key tasks of video-language understanding and discuss their common challenges: intra-modal and cross-modal interaction, cross-domain adaptation, and data preparation.
    \item We provide a clear taxonomy of video-language understanding works from three perspectives according to the three aforementioned challenges: (1) \textit{Model architecture perspective:} we classify existing works into Pre-transformer, Transformer-based, and LLM-augmented architectures to model video-language relationship. In the latter category, we discuss recent efforts that utilize the advantages of LLMs to enhance video-language understanding. (2) \textit{Model training perspective:} we categorize training methods into Pre-training and Fine-tuning to adapt video-language representations to target downstream task. (3) \textit{Data perspective:}  we summarize existing approaches that curate video-language data and annotate them to fuel the training of video-language understanding models.
    \item Finally, we provide our prospects and propose potential directions for future research.
\end{itemize}

%% file: folders/taxonomy.tex
\tikzstyle{my-box} = [
    rectangle,
    draw=hidden-black,
    rounded corners,
    text opacity=1,
    minimum height=1.5em,
    minimum width=5em,
    inner sep=2pt,
    align=center,
    fill opacity=.5,
]

\tikzstyle{leaf} = [
    my-box,
    minimum height=1.5em,
    fill=hidden-blue!90,
    text=black,
    align=left,
    font=\normalsize,
    inner xsep=2pt,
    inner ysep=4pt,
]
\begin{figure*}
    \centering
    \resizebox{\textwidth}{!}{
        \begin{forest}
            forked edges,
            for tree = {
                grow=east,
                reversed=true,
                anchor=base west,
                parent anchor=east,
                child anchor=west,
                base=left,
                font=\tiny,
                rectangle,
                draw=hidden-black,
                rounded corners,
                align=left,
                minimum width=4em,
                edge+={darkgray, line width=1pt},
                s sep=3pt,
                inner xsep=2pt,
                inner ysep=3pt,
                line width=0.8pt,
                ver/.style={rotate=90, child anchor=north, parent anchor=south, anchor=center},
            },
            where level=1{text width=6.0em, font=\scriptsize}{},
            where level=2{text width=5.5em, font=\scriptsize}{},
            where level=3{text width=19.3em, font=\scriptsize}{},
            where level=4{text width=19.3em, font=\scriptsize}{},
            where level=5{text width=19.3em, font=\scriptsize}{},
            [
                Video-Language Understanding, ver, font=\scriptsize
                [
                    Video-language \\ understanding tasks
                    [
                        Text-video retrieval
                        [
                        e.g. \citep{jiang2022cross, jin2023diffusionret, dong2022reading, pei2023clipping, lin2022eclipse, zhang2023multimodal}, text width=33em, fill=hidden-blue!90
                        ]
                    ]
                    [
                        Video captioning
                        [
                        e.g. \citep{seo2022end, wu2021weakly, zhang2020object, pan2020spatio, xu2020deep, lin2020semi}, text width=33em, fill=hidden-blue!90
                        ]
                    ]
                    [
                        VideoQA
                        [
                        e.g. \citep{xiao2023contrastive, xiao2022video, park2021bridge, li2023discovering, guo2021multi, peng2021progressive, zhao2017videoa}, text width=33em, fill=hidden-blue!90
                        ]
                    ]
                    [
                        Other tasks
                        [
                        e.g. \citep{liu2022umt, zeng2022moment, yang2021deconfounded, li2023momentdiff, lin2023learning, hwang2023meta}, text width=33em, fill=hidden-blue!90
                        ]
                    ]
                ]
                [
                    Video-language \\ understanding methods
                    [
                        Model Architecture
                        [
                            Pre-transformer, text width=5em
                            [
                            e.g. \citep{ye2017video, feichtenhofer_CVPR2016_fusion, yang2017tensor, zhao2017two}, text width=26.4em, fill=hidden-blue!90
                            ]
                        ]
                        [
                            Transformer-based, text width=5em
                            [
                            e.g. \citep{akbari2021vatt, tang2021decembert,li2023lavender, luo2022clip4clip, xue2022clip}, text width=26.4em, fill=hidden-blue!90
                            ]
                        ]
                        [
                            LLM-augmented, text width=5em
                            [
                            e.g. \citep{zhang2023video, li2023videochat, chen2023videollm, li2023llama, pan2023retrieving}, text width=26.4em, fill=hidden-blue!90
                            ]
                        ]
                    ]
                    [
                        Model Training
                        [
                            Pre-training, text width=5em
                            [
                            e.g. \citep{cheng2023vindlu, lei2021less, fu2023empirical, gao2021clip2tv, bain2021frozen}, text width=26.4em, fill=hidden-blue!90
                            ]
                        ]
                        [
                            Fine-tuning, text width=5em
                            [
                            e.g. \citep{xu2019multilevel, anne2017localizing, pan2022st, yang2022zero}, text width=26.4em, fill=hidden-blue!90
                            ]
                        ]
                    ]
                    [
                        Data perspective
                        [
                            Data curation, text width=5em
                            [
                                Manual collection, text width=5.6em
                                [
                                e.g. \citep{xue2022advancing, zellers2021merlot, castro2022fiber}, fill=hidden-blue!90, text width=19.2em,
                                ]
                            ]
                            [
                                Data augmentation, text width=5.6em
                                [
                                e.g. \citep{xing2023svformer, jiang2022semi, wang2021self}, fill=hidden-blue!90, text width=19.2em,
                                ]
                            ]
                        ]
                        [
                            Label annotation, text width=5em
                            [
                                Manual annotation, text width=5.6em
                                [
                                e.g. \citep{li2022representation, xiao2021next, castro2022wild}, fill=hidden-blue!90, text width=19.2em
                                ]
                            ]
                            [
                                Automatic generation, text width=5.6em
                                [
                                e.g. \citep{zhao2023learning, yang2023vidchapters, ventura2023covr}, fill=hidden-blue!90, text width=19.2em
                                ]
                            ]
                        ]
                    ]   
                ]
            ]
        \end{forest}
    }
    \caption{Taxonomy of Video-language Understanding}
\end{figure*}
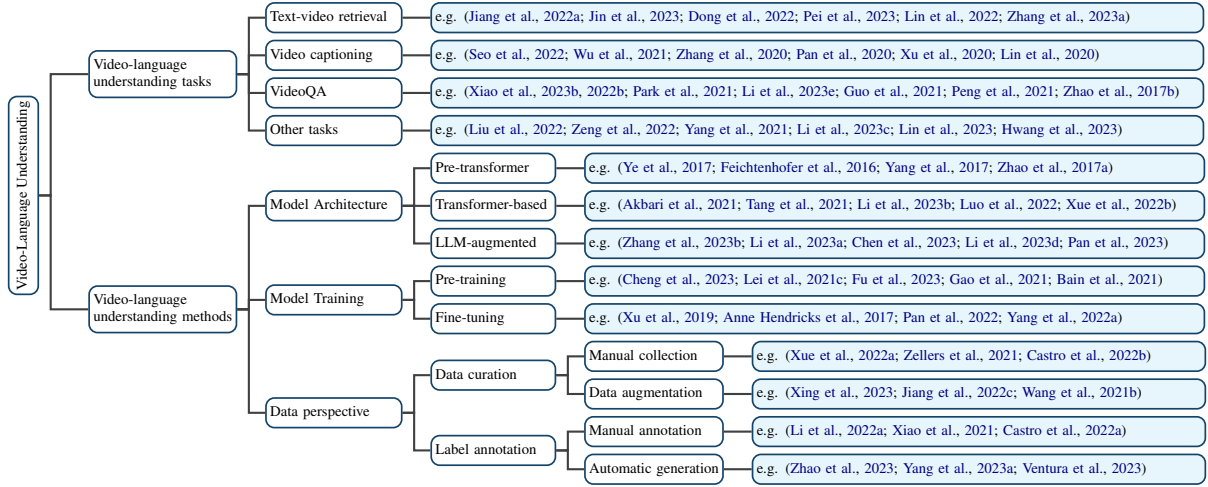

%% file: folders/02_tasks.tex
\section{Video-Language Tasks}
\label{sect:video_language_tasks}

\begin{figure*}[h!]
    \centering
    \includegraphics[width=\linewidth]{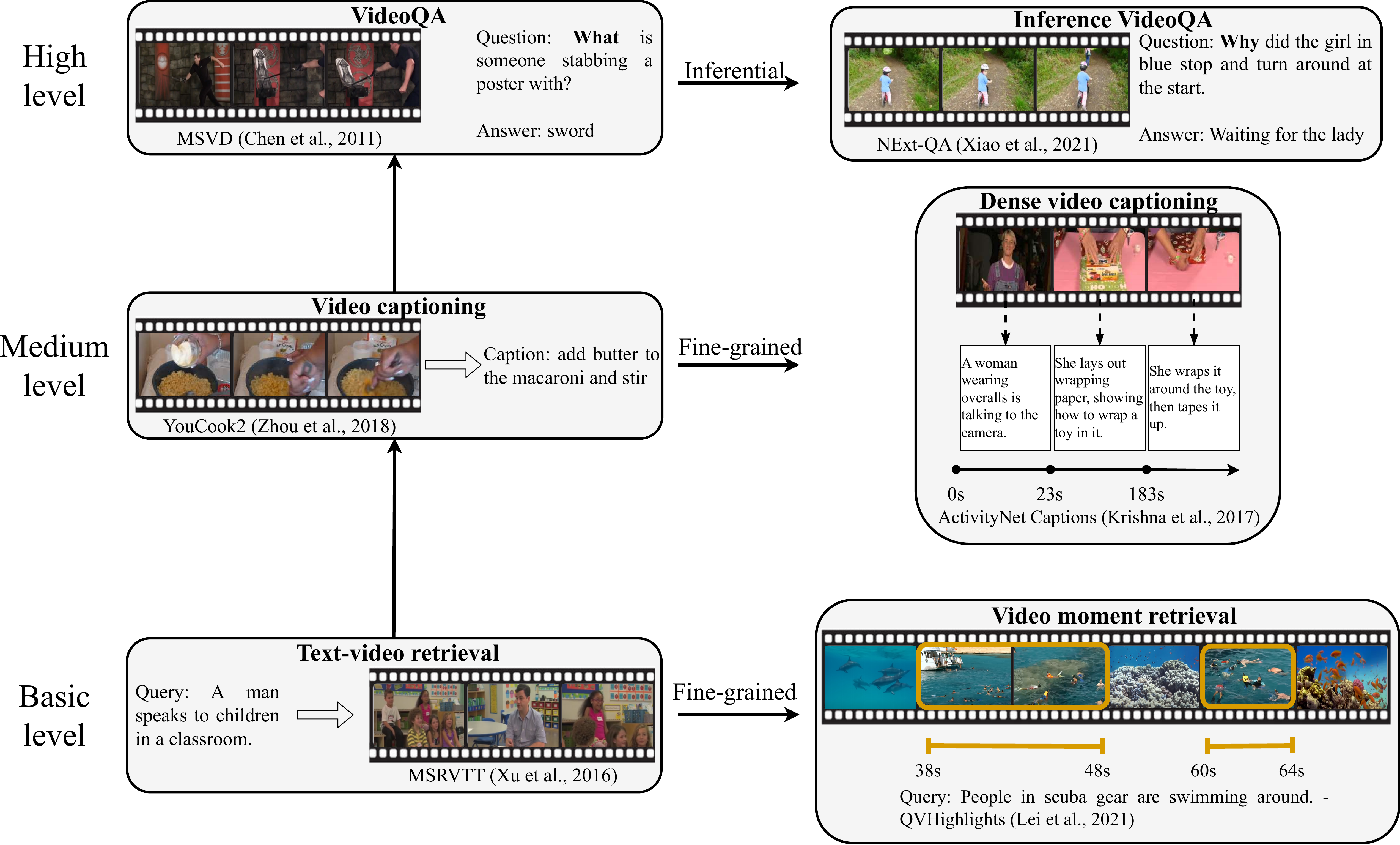}
    \caption{Level hierarchy of video-language understanding tasks.}
    \label{fig:task_hierarchy}
\end{figure*}

\noindent\textbf{Text-video retrieval.} Text-video retrieval is the task to search for the corresponding video given a language query (text-to-video), or oppositely search for the language description given a video (video-to-text).
In practical applications, returning an entire video may not be desirable. Hence, video moment retrieval (VMR) has emerged with an aim to accurately locating relevant moments within a video based on user queries. VMR examines more nuanced and fine-grained understanding to capture different concepts and events in a video in order to pinpoint specific moments rather than capturing the overall theme in standard text-video retrieval.


\noindent\textbf{Video captioning.}
Video captioning is the task to generate a concise language description for a video. A video captioning model receives as input a video and optionally a language transcript transcribed from the audio in the video. Typically, a model produces a sentence-level caption for the whole video, or might also generate a paragraph as a more detailed summary.

\noindent\textbf{Video question answering (videoQA). }Video question answering is the task to predict the correct answer based on a question $q$ and a video $v$. There are two fundamental types of VideoQA, \textit{i.e.} \textbf{multi-choice} VideoQA and \textbf{open-ended} VideoQA. In multi-choice VideoQA, a model is presented with a certain number of candidate answers and it will choose the correct answer among them. Open-ended VideoQA can be formulated as a classification problem, a generation problem, or a regression problem. Classification-based VideoQA associates a video-question pair with an answer from a pre-defined vocabulary set. Generation-based VideoQA is not restricted to a vocabulary set, in which a model can generate a sequence of tokens that represent the answer to a question. Regression-based VideoQA is often used for counting questions, \textit{e.g.} counting the repetitions of an action or counting the number of an object in a video.  

\noindent\textbf{Connections among video-language understanding tasks.}
These tasks form the three fundamental testbeds for video-language understanding capacity (see Appendix  \ref{app:task_examples} for their examples). In Figure \ref{fig:task_hierarchy}, we provide a hierarchy that describes the level-up of their video-language understanding degree. At the basic level, text-video retrieval globally associates a whole video with a textual content. In medium level, video captioning is more difficult than retrieval tasks since it needs to selectively maps entities and events within a video to the language modality. At the highest level, videoQA explores the relation of video and language content to produce the appropriate output. Each level of video-language understanding tasks is associated with a corresponding version that demands a more inferential or fine-grained understanding, \textit{e.g.} inference videoQA \citep{xiao2021next, li2022representation} with videoQA, dense video captioning \citep{zhou2018end} or video chapter generation \citep{yang2023vid2seq} with video captioning, and video moment retrieval (temporal grounding) with text-video retrieval. These more inferential or fine-grained tasks pose more challenges and play an increasingly significant role in current research heading towards the core of human intelligence \citep{fei2022searching}.

\begin{figure*}[t]
    \centering
    \includegraphics[width=\linewidth]{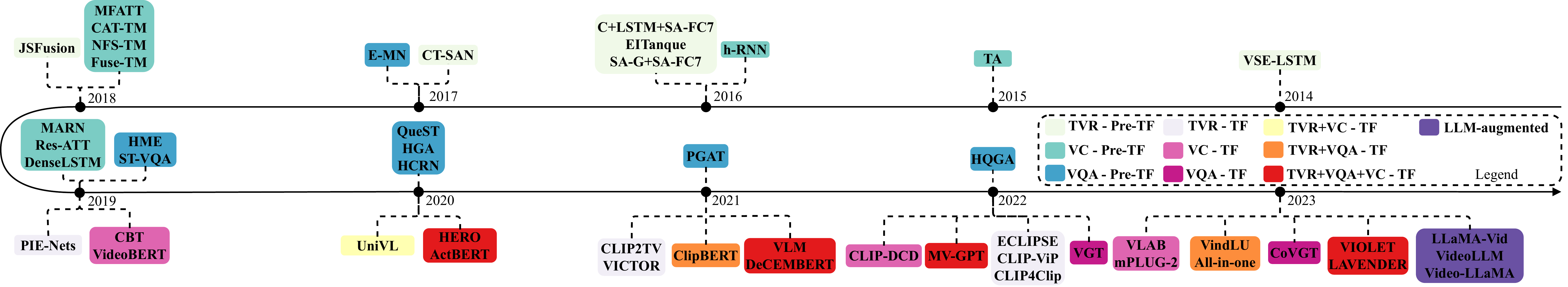}
    \caption{Timeline of the established video-language understanding methods (TVR: Text-video retrieval, VC: video captioning, VQA: video question answering, TF: Transformer, LLM: large language model). From left to right, our legend table follows the order: pre-Transformer (Pre-TF), task-specific Transformer, multi-task Transformer, and LLM-augmented architectures.}
    \label{fig:timeline}
    \vspace{-15pt}
\end{figure*}

%% file: folders/03_challenges.tex
\section{Challenges of Video-Language Understanding}
The discussed video-language understanding tasks present unique challenges compared with image-language understanding, since a video incorporates an additional temporal channel. We summarize their important challenges as follows:

\noindent\textbf{Intra-modal and cross-modal interaction.} While intra-modal interaction modeling within language can be directly taken from image-language understanding, intra-modal interaction modeling within video is different since it jointly consists of spatial interaction and temporal interaction. Spatial interaction delves into the relationships among pixels, patches, regions, or objects within an individual frame, whereas temporal interaction captures sequential dependencies among video frames or video segments. Longer video durations amplify the complexity of temporal modeling by necessitating the recognition of more objects and events in a higher number of video frames \citep{yu2020long, lin2022eclipse}, and reasoning their long-term dependencies \cite{zhao2018open}. Particular video domains, such as egocentric videos, also complicate temporal interaction modeling, as objects undergo drastic appearance and disappearance dynamics over time, posing challenges in capturing their relationships \citep{bansal2022my, tang2023egotracks}.

Given the larger semantic gap for video-language compared to image-language, cross-modal interaction plays a crucial role in video-language understanding. The interaction between visual and language features is pivotal for aligning the semantics of video and text query to associate them for text-video retrieval, or identifying relevant parts to answer the question and writing the caption in videoQA and video captioning, respectively. In addition, incorporating the interaction of motion and language features can mitigate the extraction of noisy information from videos \citep{ding2022language}. \citet{lin2022eclipse} also discover that the interaction between audio and language features can compactly capture information related to objects, actions, and complex events, compensating for sparsely extracted video frames.

\noindent\textbf{Cross-domain adaptation.} Given the infinitude of online videos, that our video-language understanding model will encounter testing scenarios which are identically distributed to our training data is an impractical assumption. Moreover, with the advent of LLM-augmented models that can tackle a variety video-language understanding tasks \citep{li2023videochat, li2023llama}, it is currently more advisable to train a model that can effectively adapt to multiple tasks and domains than to obtain a model which specializes in a specific understanding task. Furthermore, since a video can be considered as a sequence of images, training a model on video-text data is more computationally expensive than image-text data. Combined with the large-scale of recent video-language understanding models \citep{jiang2022cross, yang2022zero}, there is also a need to devise an efficient fine-tuning strategy to save the computational cost of fine-tuning these models. 


\noindent\textbf{Data preparation.} 
Although \citet{lei2021less} only use image-text data to train models for video-language understanding tasks, in essence, video-text data are crucial for the effectiveness of these models. In particular, compared with a static image, a video offers richer information with diverse spatial semantics with consistent temporal dynamics \citep{zhuang2023video}. As such, \citet{cheng2023vindlu} find that training on videos outperforms training on images, but jointly training on both data achieves the best performance. As additional evidence, \citet{yuan2023videoglue} shows that video-pretrained models outperform image-pretrained models in classifying motion-rich videos. However, video-text data takes up more storage cost than image-text data since a video comprises multiple images as video frames. Moreover, annotating a video is also more time-consuming and labor-intensive than annotating an image \citep{xing2023svformer}. Therefore, video-language understanding models have been limited by the small size of clean paired video-text corpora in contrast to billion-scale image-text datasets \citep{zhao2023learning}. Various efforts \citep{zhao2023learning, xing2023svformer} have been put into devising efficient and economical methods to curate and label video-text data.

\noindent\textbf{Addressing challenges.} These identified challenges encompass three critical perspectives: model architecture, model training, and data preparation in the field of video-language understanding. In general, there should be a synergistic relationship among these components. Specifically, model architecture should be designed to effectively capture video-language interactions. Concurrently, model training should be tailored to enable the architecture to adapt to target domains with their captured video-language interactions. Lastly, data preparation plays a pivotal role in shaping model training, which in turn significantly impacts the development of an efficacious model architecture.

%% file: folders/04_model_architecture.tex
\section{Model Architecture for Video- Language Understanding}
\label{sect:model_architecture}
\begin{figure}[t]
    \centering
    \includegraphics[width=\linewidth]{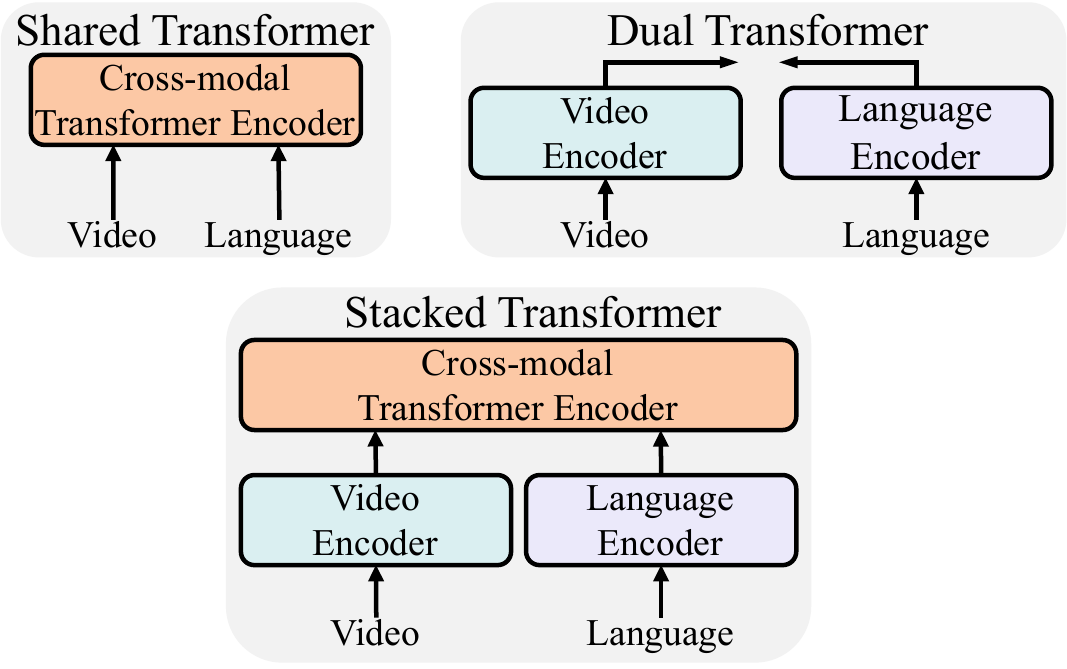}
    \caption{Illustration of video-language understanding Transformer-based architectures.
    }
    \label{fig:transformer_architectures}
    \vspace{-20pt}
\end{figure}

Addressing the challenge of intra-modal and cross-modal interaction is the key aim in designing video-language understanding model architectures, which can be divided into \textbf{Pre-transformer} and \textbf{Transformer-based architectures}. The advent of LLMs with remarkable zero-shot capability in addressing multiple tasks led to the design of \textbf{LLM-augmented architectures} that exhibit cross-domain adaptation ability to various video-language understanding tasks.

\subsection{Pre-Transformer Architecture}
Pre-transformer architectures typically comprise unimodal video and language encoders for implementing intra-modal interactions and cross-modal encoders for cross-modal interactions. 

\noindent\textbf{Unimodal encoders.} A video encoder often encodes raw videos by extracting frame appearance and clip motion features as spatial and temporal representations, respectively. As each video frame can be considered as a single image, various works have utilized CNNs to extract spatial representations \citep{simonyan_NIPS2014_twoStream, feichtenhofer_CVPR2016_fusion, zhao2017videoa}. For temporal representations, the sequential nature of RNN makes it a popular choice in pre-transformer architectures \citep{yang2017tensor, zhao2017two, venugopalan2015sequence,wang2019holistic}. Furthermore, 3D CNNs with an additional temporal channel inserted to 2D CNN have also demonstrated effectiveness in extracting spatio-temporal representations \citep{tran2017convnet, carreira2017quo}. In addition to CNN and RNN, \citet{chen2018probabilistic}, \citet{gay2019visual}, and \citet{wei2017graph} also build graphs to incorporate intra-modal relationships among video entities such as video segments or visual objects. These graph-structured works emphasize the reasoning ability of the model architecture.

A common framework of language encoder is to extract pre-trained word embeddings such as word2vec \citep{kaufman2016temporal, yu2017end} or GloVe \citep{torabi2016learning, kiros2014unifying}, then proceed with RNN-based modules such as LSTM or GRU. Such framework is taken from language model architectures before the era of Transformer.



\noindent\textbf{Cross-modal encoders.} \citet{gao2017tall} and \citet{zeng2017leveraging} apply element-wise multiplication to fuse the global video and question representations for video question answering. It demonstrates the advantage of a simple operation for video-language fusion. Attention has also been used to model video-language relations, in order to identify salient parts in video and language sentence \citep{yuan2019find}, or to refine the representation of the video based on the language question \citep{xu2017video}. Pre-transformer video-language works have also combined attention with a wide variety of techniques, including hierarchical learning \citep{baraldi2017hierarchical}, memory networks \citep{fan2019heterogeneous}, and graph networks \citep{xiao2022hqga, wei2023multi}. 

\subsection{Transformer-based Architecture}
Developed based on the self-attention mechanism, which exhaustively correlates every pair of input tokens with each other, Transformer-based architecture has the capacity to capture long-term dependencies and learn from web-scale data. It has demonstrated remarkable performance in many video-language tasks. Similar to the pre-transformer architecture, the Transformer-based framework also comprises unimodal encoders and cross-modal encoders to model intra-modal and cross-modal interactions, respectively. For unimodal encoders, several works find vision transformer for video encoding and BERT encoder for language encoding outperform RNN- and CNN-based encoding \citep{fu2021violet, bain2021frozen, seo2022end}. We then summarize fundamental types of Transformer-based architectures and illustrate them in Figure \ref{fig:transformer_architectures}.

\noindent\textbf{Shared Transformer.} Motivated by the success of Transformer in language modeling \citep{devlin2018bert}, \citet{akbari2021vatt} and \citet{wang2023all} construct a shared Transformer encoder for video-language understanding. Their encoder architectures receive the concatenation of visual patches and language tokens, then jointly calculate their interactions in a BERT-based manner. \citet{akbari2021vatt} additionally incorporate modality embeddings which comprise three values to denote three kinds of input modalities, \textit{i.e.} (video, audio, text).

\noindent\textbf{Stacked Transformer.} \citet{li2020hero} reveals that a shared Transformer encoder is weak in modeling temporal relations between videos and texts. To address this problem, they introduce a stacked Transformer architecture, with a hierarchical stack consisting of unimodal encoders to encode video and language inputs separately, and then a cross-modal Transformer to compute video-language interactions. A multitude of video-language understanding works follow such design to stack a cross-modal Transformer-based encoder above unimodal encoders \citep{fu2023empirical, li2023lavender, lei2021less, wei2022audio, luo2022clip4clip, nie2022search, wei2024learning}. To perform video captioning, \citet{seo2022end} and \citet{luo2020univl} further insert a causal Transformer-based decoder that generates language tokens based on the encoded cross-modal representations.

\noindent\textbf{Dual Transformer.} Dual Transformer architectures have been favored for text-video retrieval \citep{luo2022clip4clip, bain2021frozen, bain2022clip, lin2022eclipse, xue2022clip}. These architectures use two Transformer encoders to encode video and language separately, yielding global representations for each input modality, then applying simple operations such as cosine similarity to compute cross-modal interaction. Such a separate encoding scheme enables them to mitigate the computational cost of computing pairwise interactions between every pair of video and language inputs. 
They have accomplished not only efficiency but also effectiveness in text-video retrieval problems.

\subsection{LLM-Augmented Architecture}
\label{sect:llm_augmented_architecture}



Large language models (LLMs) have achieved impressive results in simultaneously tackling multiple NLP tasks. Recent efforts have sought to apply LLMs for video-language understanding to extend its cross-domain adaptation ability to video-language settings \citep{chen2023videollm, li2023videochat, nguyen2025temporal}. These efforts can be categorized into two approaches. The first approach employs LLM as a controller and video-language understanding models as helping tools. The controller will call the specific tool according to the language input instruction. The second approach utilizes LLM as the output generator and seeks to align video pre-trained models to the LLM. For video-language understanding, since the second approach dominates the first one with a long list of recent works \citep{chen2023videollm, li2023videochat, chen2023videollm, li2023llama, zhang2023video, maaz2023video}, we review them as follows:

\noindent\textbf{LLM as Output Generator.} The framework comprises a visual encoder, a semantic translator, and an LLM as the output generator. Regarding visual encoder, LLM-augmented architectures often use vision transformer and CNN models of the pre-Transformer and Transformer-based architectures \citep{chen2023videollm}. Since an LLM has never seen a video during its training, a semantic translator is needed to translate the visual semantics of a video to the LLM’s semantics. For the translator, Video-LLaMA \citep{zhang2023video} and VideoChat \citep{li2023videochat} implement a Q-Former as a Transformer-based module that uses a sequence of query embeddings that interact with visual features of the video to extract informative video information. Instead of Q-Former, VideoLLM \citep{chen2023videollm}, Video-ChatGPT \citep{maaz2023video}, and LLaMA-Vid \citep{li2023llama} find that a simple linear projection that projects visual features into the LLM’s input dimension can achieve effective performance. Subsequently, these visual-based query embeddings or projected visual features are combined with the language instruction to become the input fed to the LLM to produce the final output. 

\subsection{Architecture Analysis}
\label{sect:architecture_analysis}
In Figure \ref{fig:timeline}, we show the timeline of video-language understanding methodologies, categorized according to our defined architecture taxonomy and their affiliated downstream tasks. The evolution of pre-transformer models aligns with our hierarchy of video-language understanding levels, \textit{i.e.} models for video captioning generally appear subsequent to those for text-video retrieval, followed by the development of videoQA models. Owing to their impressive capacity, Transformer-based models capable of addressing multiple tasks have been introduced concurrently with task-specific Transformer frameworks. Recently, large language models (LLMs) have gained prominence for their superior in-context learning ability, enabling them to handle diverse tasks without fine-tuning. Consequently, new LLM-augmented architectures have emerged to utilize this capability to address multiple understanding tasks.

Among Transformer-based architectures, the dual Transformer stands out as the most effective for text-video retrieval, adeptly associating global semantics of video and language modality. On the other hand, the stacked Transformer architecture excels at facilitating intra-modal and cross-modal interactions through its specialized unimodal and cross-modal encoders. These encoders are particularly efficient at correlating video content with the question in videoQA. Additionally, for video captioning, cross-modal encoder plays a crucial role in translating video content into textual descriptions. Recent LLM-augmented models have begun to outperform Transformer-based architectures in videoQA, signalling their potential as the next frontier in video-language understanding research. We provide full details of performance in text-video retrieval, video captioning, and videoQA tasks in Table \ref{tab:app_exp_text_video_retrieval}, \ref{tab:app_exp_video_captioning}, and \ref{tab:app_exp_videoqa}, respectively.

%% file: folders/05_model_training.tex
\section{Model Training for Video-Language Understanding}
\label{sect:model_training}
Model training seeks to address the cross-domain adaptation ability of video-language understanding models. To achieve this goal, pre-training strategies have been devised to gain world knowledge that generalizes across multiple scenarios, then task-specific fine-tuning is conducted to specifically improve downstream task performance.
\subsection{Pre-training for Video-Language Understanding}
In this section, we mainly summarize pre-training strategies for video-language understanding models into three groups:

\noindent\textbf{Language-based pre-training.} The most popular language-based pre-training task is masked language modeling (MLM) \citep{lei2021less, sun2019videobert, cheng2023vindlu}, which randomly masks a portion of words in the language input and trains the model to predict the masked words based on unmasked language words and video entities. Instead of masking a portion of words, UniVL \citep{luo2020univl} and VICTOR \citep{lei2021understanding} discover that masking the whole language modality benefits video captioning task. MLM can be combined with other language-based pre-training task, \textit{e.g.} masked sentence order modeling which is to classify the original order of the shuffled language sentences \citep{lei2021understanding}.

\noindent\textbf{Video-based pre-training.} Video-based pre-training tasks help video-language models capture contextual information in the video modality. As a counterpart of MLM, masked video modeling (MVM) trains the model to predict the portion of masked video entities based upon the unmasked entities and language words. The continuous nature of videos leads to different choices of video entities, such as frame patches \citep{li2020hero} or video frames \citep{fu2021violet}. In terms of the training objective, \citet{li2020hero} use L2 regression loss to train the model to predict pre-trained features of the masked video frames extracted by ResNet and SlowFast models, while \citet{fu2021violet} use cross-entropy loss to train the model to predict the masked visual tokens, which are quantized by a variational autoencoder from visual frame patches. 

\noindent\textbf{Video-text pre-training.} Video-text pre-training is crucial for a model to capture video-language relation. \citet{xue2022clip}, \citet{gao2021clip2tv}, and \citet{bain2021frozen} utilize a framework of video-text contrastive learning to produce close representations for semantically similar video and language inputs. These works focus on creating a joint semantic space that aligns separate representations of video and language. Instead of separate representations, \citet{tang2021decembert}, \citet{fu2021violet}, and \citet{li2023lavender} enable video and textual representations to interact with each other and use a single token to represent the cross-modal input, which is forwarded to predict whether the video-text pair is matched or not. In these two pre-training frameworks, not only video-text data but also image-text data are utilized during pre-training, in which an image is considered as a video with a single frame.

Contrastive learning has revealed promising results  \citep{lin2022eclipse, gao2021clip2tv, xue2022clip, nguyen2022adaptive, nguyen2021contrastive, nguyen2024topic, nguyen2024kdmcse, nguyen2023improving, wu2023infoctm, wu2024dynamic, nguyen2025multi, nguyen2025motion}. MLM has contributed to enhancing VideoQA since the task resembles MLM in predicting the language word given a video-language pair (the question is the language input in videoQA). Compared to these pre-training strategies, MVM does provide performance gain for video-language understanding but its gain is less significant. For more details about pre-training, please refer to \citep{cheng2023vindlu}.

\vspace{-2mm}
\subsection{Fine-tuning for Video-Language Understanding}
Task-specific fine-tuning is commonly used by pre-Transformer architectures to train from scratch since these models do not have sufficient parameter capacity to learn generalizable features through pre-training. It is also widely adopted by Transformer-based architectures to improve the performance for a specific downstream task. Moreover, LLM-augmented architectures also utilize instruction tuning as a variant of fine-tuning, to adapt from the visual and audio spaces to the LLM language space.

\noindent\textbf{Fine-tuning strategies.} Normally, all of the model parameters are updated during fine-tuning \citep{gao2017tall, xu2019multilevel, anne2017localizing, nguyen2023demaformer, wu2023effective}. However, in cases computational resources or training data are limited, only adaptation layers such as low-rank adapters \citep{pan2022st, yang2022zero, nguyen2024read} or learnable prompt vectors \citep{ju2022prompting} are fine-tuned to reduce training cost or prevent overfitting. Such risks also apply for LLM-augmented architectures discussed in Section \ref{sect:llm_augmented_architecture}, since LLMs exhibit a billion scale of parameters, thus incurring excessively huge cost if full fine-tuning is conducted. For such models, \citet{zhang2023video} and \citet{li2023llama} design a two-stage instruction tuning strategy which only fine-tunes the semantic translator. The first stage trains the model to generate the textual description based on the combined video and the language instruction, in order to align visual representations extracted by the visual encoder with the language space of LLM. The second stage is often performed on small-scale video-text pairs manually collected by the authors to further tailor the output features of the translator towards the target domains.


%% file: folders/06_data.tex
\section{Data Perspective for Video-Language Understanding}
\label{sect:data}
In this section, we analyze data preparation approaches for video-language understanding models, and provide details of the datasets in Appendix \ref{app:details_dataset}.
\subsection{Data curation}

\noindent\textbf{Manual collection.} To curate video-language data, multiple works search for publicly available online videos, which exhibit a wide diversity of content. Video-language datasets with online videos are mostly aimed for pre-training models to learn generalizable knowledge, \textit{e.g.} HowTo100M \citep{miech2019howto100m} and YT-Temporal-180M \citep{zellers2021merlot}, or they can also be used for fine-tuning, \textit{e.g.} MSRVTT \citep{xu2016msr} and YouCook2 \citep{zhou2018towards}. To satisfy a certain requirement, videos different from the online ones can be inherited from existing datasets, \textit{e.g} \citet{xiao2021next} utilize 6,000 videos from VidOR dataset and \citep{li2022representation} inherit 546,882 videos from Kinetics-700 since they describe scenes of daily life and real world, respectively. Apart from making use of existing datasets' and online videos, videos can also be recorded by human annotators to enable quality control \citep{sthsth, damen2022rescaling}.

\noindent\textbf{Data augmentation.} Rather than manually collecting videos from external sources, \citet{xing2023svformer} and \citet{jiang2022semi} explore data augmentation techniques which are particularly designed for videos. In detail, their TubeTokenMix mixes two videos in which the mixing coefficient is defined upon the temporal dimension, and their temporal shift randomly shifts video frame features backward or forward over the temporal dimension. These techniques outperform standard augmentation approaches for image data, such as CutMix \citep{yun2019cutmix}, Mixup \citep{zhang2017mixup}, and PixMix \citep{hendrycks2022pixmix}.

\subsection{Label annotation}
\noindent\textbf{Manual annotation.} Several works \citep{li2022representation, lei2021detecting, xiao2021next} use human annotators since they provide high-quality labels. However, such approach is expensive, particularly when dealing with video data. For example, annotating QVHighlights dataset \citep{lei2021detecting} costs approximately \$16,000 for 10K videos and 3 months to complete. Similarly, NExT-QA \citep{xiao2021next} needs 100 undergraduate students and 1 year to annotate only 5K videos.

\noindent\textbf{Automatic generation.} Directly taking language transcripts of YouTube videos as textual labels could reduce annotation cost \citep{miech2019howto100m, xue2022advancing, zellers2021merlot}. However, these labels have been shown to be grammatically incorrect and temporally misalign with the video content \citep{tang2021decembert, nguyen2024meta}. Motivated by the success of LLMs, \citet{zhao2023learning} train a system consisting of a TimeSformer-L visual encoder and a GPT-2XL decoder to write dense captions for videos. Moreover, \citet{li2023videochat} use GPT-4 to generate summaries for movie synopses. 

%% file: folders/07_future_directions.tex
\section{Future Directions}
\label{sect:future_directions}

\noindent\textbf{Fine-grained understanding.} Existing methods excel at video-language understanding at a coarse-grained level, enabling effective responses to questions like ``\textit{what is}'' or the recognition of global events without significant difficulty \citep{xiao2021next}. Nevertheless, limiting comprehension to this coarse level could hinder practical utility of existing systems. In real-world scenarios, a user might require a precise timestamp and location of an object within a video \citep{jiang2022video}, or request the AI agent to forecast potential alternative events, which is a common need in predictive analytics \citep{xiao2021next, li2022representation}. These tasks necessitate an advanced understanding and inference capability regarding the causal and temporal relationships present in a video. At present, models exhibit a constrained visio-linguistic capacity to engage in temporal reasoning, categorizing them as image-sequence-and-language models rather than video-language models \citep{kesen2023vilma}. Therefore, future research in this direction deserves more attention and exploration. 

\noindent\textbf{Long-form video-language understanding.} Current understanding systems have demonstrated remarkable performance on short video clips lasting several seconds. However, they tend to struggle when switching to long-form videos which last several minutes or hours. To enhance the applicability of these systems, it is essential to enhance their capability of understanding long-form videos. Current approaches mainly feature reducing computational cost through architectures more efficient than Transformer-based ones such as state space models \citep{yang2024vivim, li2024videomamba, nguyen2024encoding}, which can be considered as linear RNN with specifically designed fixed weights, or compensating sparsely extracted video frames with additional information \citep{lin2022eclipse}. In general, how to effectively model long-form videos and adapt them to the joint context with language deserves more attention.


\noindent\textbf{Trustworthiness of video-language understanding models.} Although modern video-language understanding systems have demonstrated remarkable performance, their black-box nature undermines our trust to deploy them. In particular, we still do not precisely understand what part of the video a videoQA model looks at to answer the question \citep{li2022equivariant}, or how video and language semantic information flows into the common representation space of the video retrieval model \citep{jia2022adversarial}. Furthermore, adversarial noise sensitivity or hallucination of video-language understanding models are also open problems. Future trustworthiness benchmarks such as \citep{xiao2023can,wang2021dutrust} for video-language understanding are of great significance towards practical systems.

%% file: folders/08_conclusion.tex
\section{Conclusion}
In this paper, we survey the broad research field of video-language understanding. Particularly, we categorize related video-language understanding tasks and discuss meaningful insights from model architecture, model training, and data perspectives. We thoroughly analyze each perspective, and finally conclude with promising future directions. We hope our survey can foster more research towards constructing effective AI systems that can comprehensively understand dynamic visual world and meaningfully interact with humans.

%% file: folders/09_limitations.tex
\section{Limitations}
Although we have sought to comprehensively analyze the literature of video-language understanding, we might not fully cover all of the tasks, model architectures, model training, and data perspectives. Therefore, we complement the survey with a repository \url{https://github.com/nguyentthong/video-language-understanding}. The repository comprises the latest papers, datasets, and their open-source implementations. We will periodically update the repository to trace the progress of the latest research. 

%% file: folders/10_acknowledgement.tex
\section{Acknowledgement}
This research/project is supported by the National Research Foundation, Singapore under its AI Singapore Programme (AISG Award No: AISG2-TC-2022-005).

%% file: folders/appendix.tex
\onecolumn
\part{Appendix}
\section{Examples of Video-Language Understanding tasks}
\label{app:task_examples}
\begin{figure*}[h!]
    \centering
    \begin{subfigure}[t]{0.3\linewidth}
        \centering
        \includegraphics[width=\linewidth]{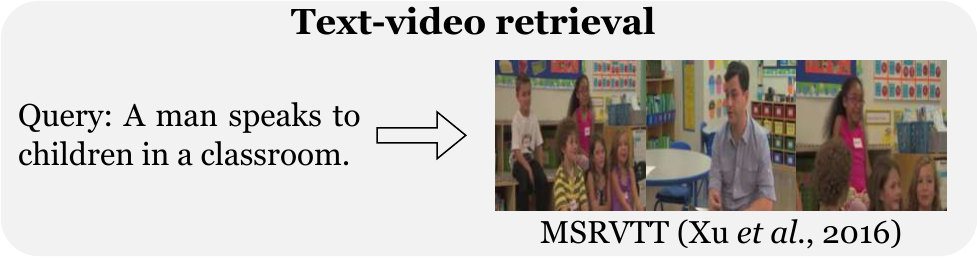}
    \end{subfigure}
    \begin{subfigure}[t]{0.3\linewidth}
        \centering
        \includegraphics[width=\linewidth]{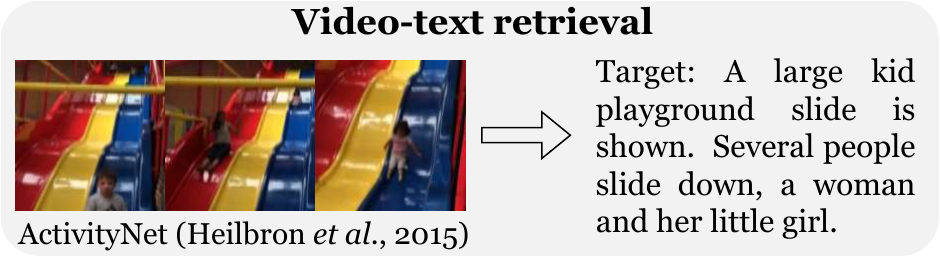}
    \end{subfigure}
    \begin{subfigure}[t]{0.3\linewidth}
        \centering
        \includegraphics[width=\linewidth]{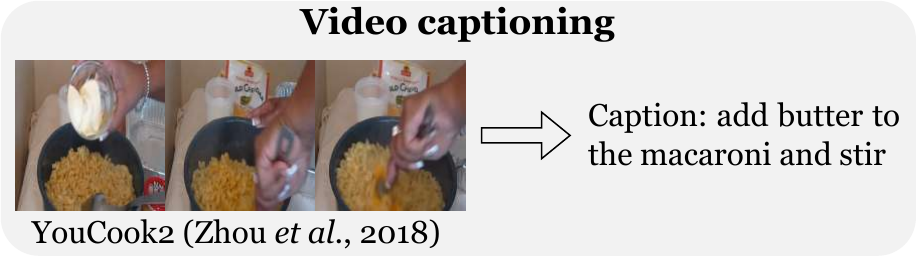}
    \end{subfigure}\\
    \begin{subfigure}[t]{0.3\linewidth}
        \centering
        \includegraphics[width=\linewidth]{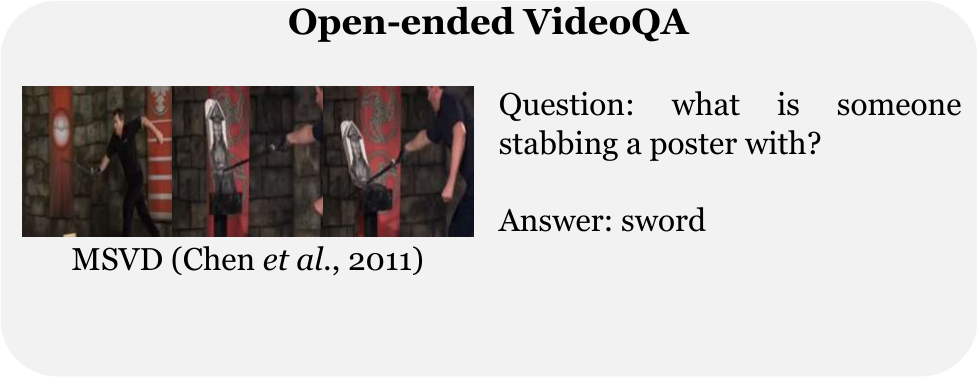}
    \end{subfigure}
    \begin{subfigure}[t]{0.3\linewidth}
        \centering
        \includegraphics[width=\linewidth]{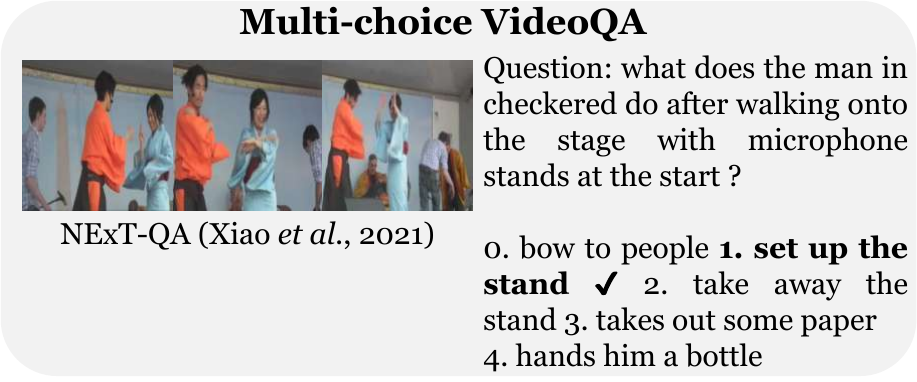}
    \end{subfigure}
    \begin{subfigure}[t]{0.3\linewidth}
        \centering
        \includegraphics[width=\linewidth]{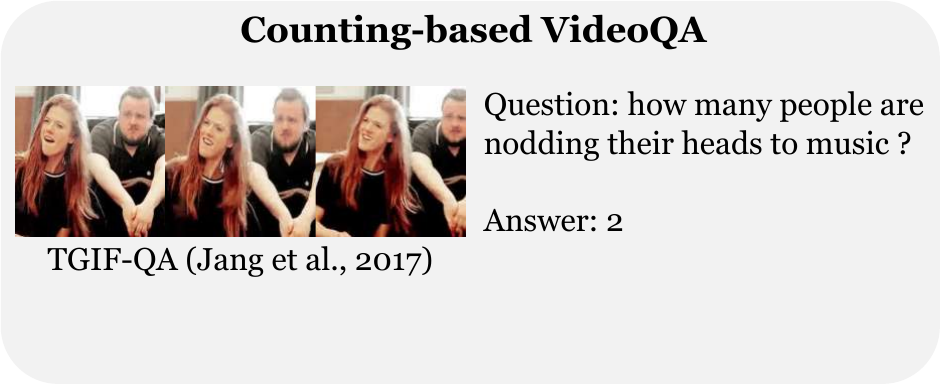}
    \end{subfigure}
    \caption{Illustration of text-video retrieval, video captioning, and video question answer (videoQA) tasks. 
    }
    \label{fig:illustration_video_language_understanding_tasks}
\end{figure*}

\begin{figure*}[h!]
    \centering
    \begin{subfigure}[t]{0.32\linewidth}
        \centering
        \includegraphics[width=\linewidth]{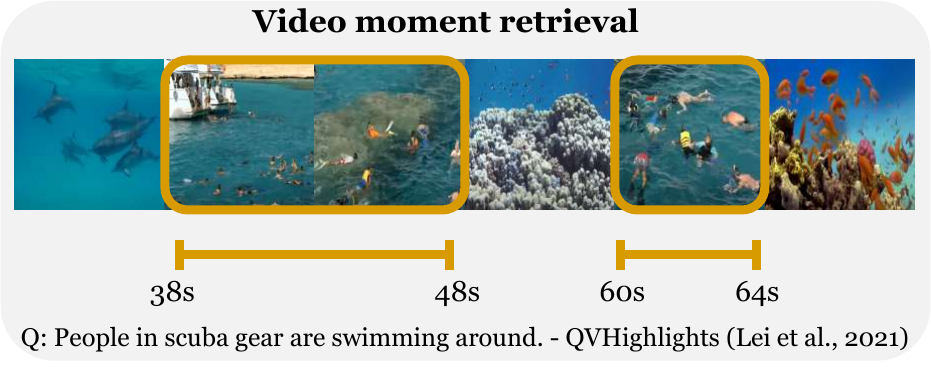}
    \end{subfigure}
    \begin{subfigure}[t]{0.29\linewidth}
        \centering
        \includegraphics[width=\linewidth]{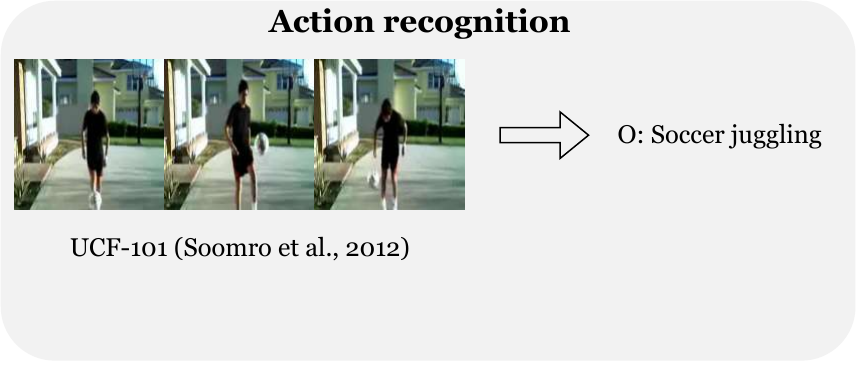}
    \end{subfigure}
    \begin{subfigure}[t]{0.32\linewidth}
        \centering
        \includegraphics[width=\linewidth]{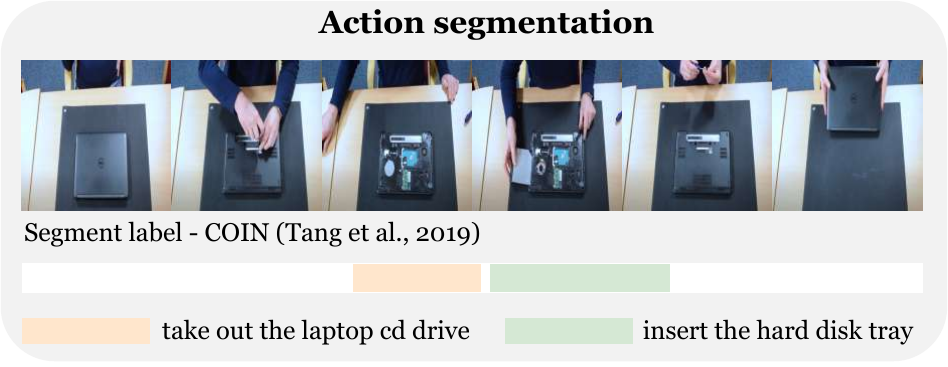}
    \end{subfigure} \\
    \begin{subfigure}[t]{0.17\linewidth}
        \centering
        \includegraphics[width=\linewidth]{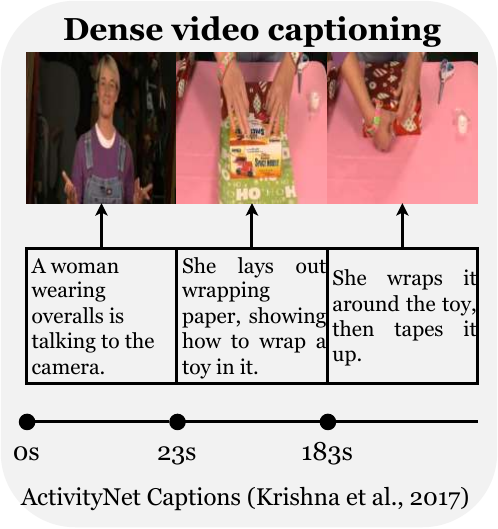}
    \end{subfigure}
    \begin{subfigure}[t]{0.33\linewidth}
        \centering
        \includegraphics[width=\linewidth]{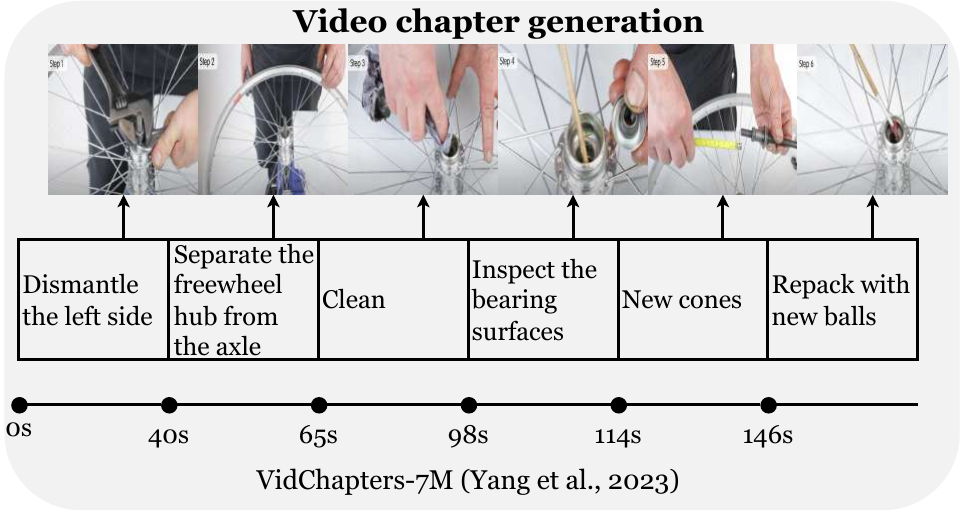}
    \end{subfigure}
    \begin{subfigure}[t]{0.45\linewidth}
        \centering
        \includegraphics[width=\linewidth]{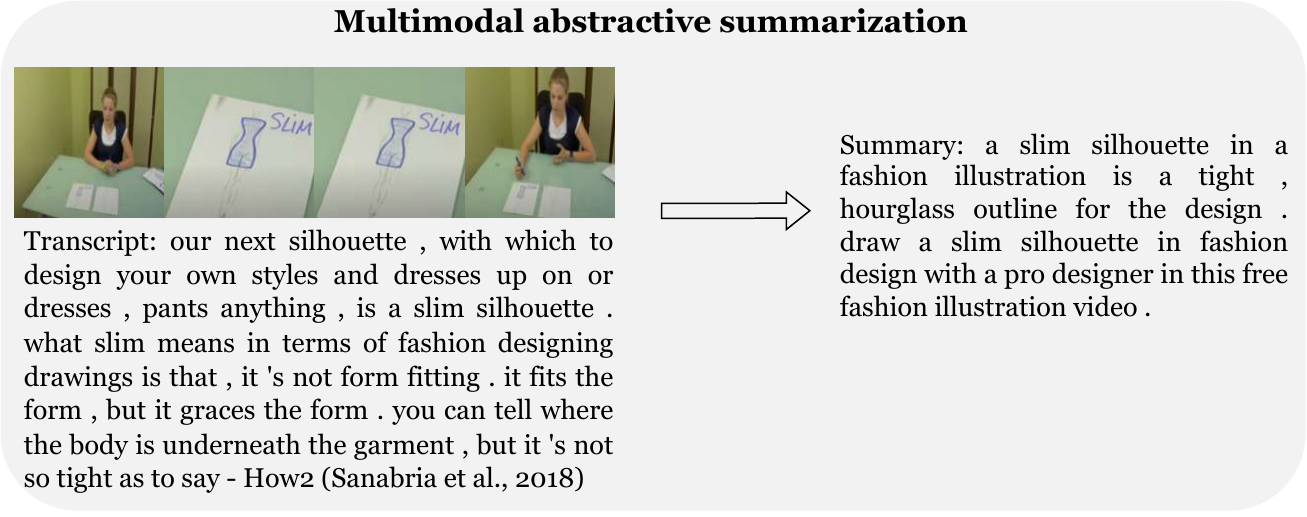}
    \end{subfigure}
    \caption{More illustration of video moment retrieval, action recognition, action segmentation, dense video captioning, video chapter generation, and multimodal abstractive summarization tasks.}
    \label{fig:more_illustration_video_language_understanding_tasks}
\end{figure*}

In this appendix, we provide examples of video-language understanding tasks in Figure \ref{fig:illustration_video_language_understanding_tasks} and \ref{fig:more_illustration_video_language_understanding_tasks}.

\section{Analysis of Video-Language Understanding datasets}
\label{app:details_dataset}
Due to page limit, details of the datasets for video-language understanding tasks are listed in Table \ref{tab:datasets}. We categorize all the datasets according to their tasks that they support. While datasets for downstream tasks and fine-tuning have been consistently developed, those for pre-training emerge subsequently to the introduction of the Transformer architecture. Although pre-training and downstream video-language understanding datasets pursue different goals, they predominantly originate from the Internet. Regarding downstream datasets, more recent ones aim to present new technical challenges, such as evaluating reasoning and inference abilities \citep{xiao2021next, li2022representation}, or examining long-form modeling capacity of video-language understanding models \citep{mangalam2023egoschema}.

\begin{table*}[h]
\centering
\resizebox{\linewidth}{!}{
\begin{tabular}{l|c|c|c|ccc}
\hline
\textbf{Methods}       & \textbf{Model architecture}      & \textbf{Video}             & \textbf{Text}          & \textbf{R@1}  & \textbf{R@5}  & \textbf{R@10} \\ \hline
VSE-LSTM   \citep{kiros2014unifying}    &     \multirow{6}{*}{Pre-TF}                    & ConvNet/OxfordNet & GloVe-LSTM    & 3.8  & 12.7 & 17.1 \\
C+LSTM+SA-FC7 \citep{torabi2016learning} &                         & VGG               & GloVe-LSTM    & 4.2  & 12.9 & 19.9 \\
EITanque  \citep{kaufman2016temporal}    &                         & VGG               & word2vec-LSTM & 4.7  & 16.6 & 24.1 \\
SA-G+SA-FC7 \citep{torabi2016learning}   &                         & VGG               & GloVe         & 3.1  & 9.0  & 13.4 \\
CT-SAN    \citep{yu2017end}    &                      & RN                & word2vec-LSTM & 4.4  & 16.6 & 22.3 \\ 
JSFusion  \citep{yu2018joint}    &  & RN                & GloVe-LSTM    & 10.2 & 31.2 & 43.2 \\ \hline
All-in-one  \citep{wang2023all}   & Shared TF               & Linear               & BT            & 37.9 & 68.1 & 77.1 \\
VLM   \citep{xu2021vlm}     & Shared TF               & S3D               & BT            & 28.1 & 55.5 & 67.4 \\
DeCEMBERT   \citep{tang2021decembert}     & Shared TF               & RN              & BT            & 17.5 & 44.3 & 58.6 \\
ActBERT   \citep{zhu2020actbert}     & Stacked TF               & Faster-RCNN              & BT & 16.3 & 42.8 & 56.9 \\
VIOLET   \citep{fu2023empirical}     & Stacked TF               & VS-TF               & BT            & 37.2 & 64.8 & 75.8 \\
VindLU   \citep{cheng2023vindlu}     & Stacked TF               & ViT               & BT            & \underline{48.8} & \underline{72.4} & \underline{82.2} \\
HERO   \citep{li2020hero}       & Stacked TF                & RN+SlowFast       & BT            & 16.8 & 43.4 & 57.7 \\
MV-GPT   \citep{seo2022end}     & Stacked TF                & ViViT             & BT            & 37.3 & 65.5 & 75.1 \\
CLIP2TV  \citep{gao2021clip2tv}    & Dual TF                 & ViT               & CLIP-text     & 32.4	 & 58.2 & 68.6 \\
CLIP-ViP  \citep{xue2022advancing}    & Dual TF                 & ViT               & CLIP-text     & \textbf{49.6} & \textbf{74.5} & \textbf{84.8} \\
CLIP4Clip \citep{luo2022clip4clip}    & Dual TF                 & ViT               & CLIP-text     & 44.5 & 71.4 & 81.6 \\ \hline
\end{tabular}}
\caption{Performance on text-video retrieval. (Pre-TF: Pre-transformer, Shared TF: Shared Transformer, Stack TF: Stack Transformer, Dual TF: Dual Transformer, RN: ResNet/ResNeXt \citep{he_CVPR2016_resnet, xie_CVPR2017_resnext}, ViT: Vision Transformer \citep{dosovitskiy2020image}, BT: BERT \citep{devlin2018bert}, ViViT: Video Vision Transformer \citep{arnab2021vivit}). We report recall at rank 1 (R@1), 5 (R@5), and 10 (R@10). We choose MSRVTT as one of the most popular datasets for text-video retrieval.}
\label{tab:app_exp_text_video_retrieval}
\end{table*}

\begin{table*}[h]
\centering
\resizebox{\linewidth}{!}{
\begin{tabular}{l|c|c|ccc}
\hline
\textbf{Methods} & \textbf{Model architecture }             & \textbf{Video} & \textbf{BLEU-4} & \textbf{METEOR} & \textbf{CIDEr} \\ \hline
TA      \citep{yao2015describing}                    &          \multirow{9}{*}{Pre-TF}                 & Video: 3D-CNN             & 36.5   & 25.7  &  - \\
h-RNN     \citep{yu2016video}                  &                           & Video: VGG                & 36.8   & 25.9 &  - \\
MFATT    \citep{long2018video}                   &    & Video: RN+C3D             & 39.1   & 26.7  & - \\
CAT-TM   \citep{long2018video}      &                           & Video: RN+C3D             & 36.6   & 25.6 &  - \\
NFS-TM     \citep{long2018video}                 &                           & Video: RN+C3D             & 37.0   & 25.9 & - \\
Fuse-TM     \citep{long2018video}                &                           & Video: RN+C3D             & 37.5   & 25.9 & - \\ 
MARN     \citep{pei2019memory}                &                           & Video: RN             & -   & - & 46.8 \\ 
Res-ATT     \citep{li2019residual}                &                           & Video: RN             & 37.0   & 26.9 & 40.7 \\ 
DenseLSTM     \citep{zhu2019attention}                &                           & Video: VGG             & 38.1   & 27.2 & 42.8 \\  \hline
VIOLET   \citep{fu2023empirical}     & \multirow{8}{*}{Stacked TF}               & VS-TF               &             - & - &  58.0 \\
LAVENDER      \citep{li2023lavender}                               &                     & VS-TF                                   & -          & -     & 57.4   \\
VLAB     \citep{he2023vlab}                   &  & EVA-G                     & 54.6   & 33.4 &  74.9 \\
UniVL   \citep{luo2020univl}                    &                           & S3D                       & 41.8   & 28.9  &  50.0 \\
MV-GPT    \citep{seo2022end}                  &                           & ViViT                     & 48.9   & 38.7 &  60.0 \\
CLIP-DCD      \citep{yang2022clip}              &                           & ViT                       & 48.2   & 30.9  & 64.8 \\
DeCEMBERT    \citep{tang2021decembert}               &                           & RN                        & 45.2   & 29.7 &  52.3 \\
mPLUG-2    \citep{xu2023mplug}                 &                           & ViT                       & 57.8   & 34.9  & 80.3 \\ \hline
\end{tabular}}
\caption{Performance on video captioning. (Pre-TF: Pre-transformer, Stacked TF: Stacked Transformer, RN: ResNet/ResNeXt \citep{he_CVPR2016_resnet, xie_CVPR2017_resnext}, ViViT: Video Vision Transformer \citep{arnab2021vivit}, EVA-G: \citet{fang2023eva}). We report BLEU-4 and METEOR, which are two popular metrics for language generation. We choose MSRVTT as one of the most popular datasets for video captioning.}
\label{tab:app_exp_video_captioning}
\end{table*}

\begin{table*}[h]
\centering
\resizebox{\linewidth}{!}{
\begin{tabular}{l|c|c|c|cc}
\hline
\multicolumn{1}{c|}{\multirow{2}{*}{\textbf{Methods}}} & \multirow{2}{*}{\textbf{Architecture}} & \multirow{2}{*}{\textbf{Video}} & \multirow{2}{*}{\textbf{Text}} & \multicolumn{2}{c}{\textbf{Dataset}} \\ 
                         &                               &                        &                       & MSRVTT        & MSVD        \\ \hline
E-MN    \citep{xu2017video}                                    & \multirow{8}{*}{Pre-TF}       & VGG + C3D               & GloVe-LSTM            & 30.4          & 26.7           \\
QueST    \citep{jiang2020divide}                                    &        & RN + C3D               & GloVe-LSTM            & 40.0          & -           \\
HME      \citep{fan2019heterogeneous}                                    &                               & RN/VGG + C3D           & GloVe-GRU             & 34.6          & 36.1        \\
HGA        \citep{jiang2020reasoning}                                  &                               & RN/VGG + C3D           & GloVe-GRU             & 33.0          & 33.7        \\
ST-VQA  \citep{jang2019video}                                     &                               & RN+C3D                 & GloVe-LSTM            & 35.5          & 34.7        \\
PGAT       \citep{peng2021progressive}                                  &                               & Faster-RCNN            & GloVe-LSTM            & 38.1          & 39.0        \\
HCRN          \citep{le2020hierarchical}                               &                               & RN                     & GloVe-LSTM            & 35.6          & 36.1  
\\
HQGA          \citep{xiao2022hqga}                               &                               & Faster-RCNN                     & BERT-LSTM            & 38.6          & 41.2       
\\ \hline
All in one     \citep{wang2023all}                              & Shared TF                     & Linear                    & BT                    & 44.3          & 47.9        \\
LAVENDER      \citep{li2023lavender}                               & Stacked TF                     & VS-TF                  & BT                    & 45.0          & 56.6        \\
DeCEMBERT   \citep{tang2021decembert}     & Stacked TF               & RN               & BT             & 37.4 & - \\
VindLU   \citep{cheng2023vindlu}     & Stacked TF               & ViT               & BT             & 44.6 & - \\
VIOLET        \citep{fu2023empirical}                               & Stacked TF                     & VS-TF                  & BT                    & 44.5          & 54.7        \\
ClipBERT        \citep{lei2021less}                             & Stacked TF                     & CLIP-text              & BT                    & 37.4          & -           \\
VGT    \citep{xiao2022video}                                       & Dual TF                       & Faster-RCNN            & BT                    & 39.7          & -           \\
CoVGT     \citep{xiao2023contrastive}                                   & Dual TF                       & Faster-RCNN            & BT                    & 40.0          & -        \\ \hline
LLaMA-Vid      \citep{li2023llama}                              & LLM-Augmented                 & EVA-G                  & Vicuna                & 58.9          & 70.0    \\ \hline   
\end{tabular}}
\caption{Performance on videoQA. (Pre-TF: Pre-transformer, Dual TF: Dual Transformer, RN: ResNet/ResNeXt \citep{he_CVPR2016_resnet, xie_CVPR2017_resnext}, BT: BERT \citep{devlin2018bert}, VS-TF: Video Swin Transformer \citep{liu2021video}, EVA-G: \citet{fang2023eva}). We report accuracy of the methods. We choose MSRVTT and MSVD as two of the most popular datasets for videoQA. }
\label{tab:app_exp_videoqa}
\end{table*}

\begin{table*}[t]
\centering
\resizebox{\linewidth}{!}{
\begin{tabular}{l|l|c|c|c}
\hline\hline
\multicolumn{1}{c|}{\textbf{Dataset}}            & \multicolumn{1}{c|}{\textbf{Video source}}     & \textbf{Annotation}   & \textbf{Tasks}            & \textbf{\#Videos/\#Routes} \\ \hline
 MSVD   \citep{chen2011collecting}            & YouTube videos         & Manual       & TVR, VC, VideoQA & 1.9K              \\
 MSRVTT     \citep{xu2016msr}        & Web videos         & Manual       & TVR, VC, VideoQA & 7.2K              \\
 Ego-QA     \citep{nguyen2024encoding}        & Egocentric videos         & Manual       & VideoQA & 18.5K              \\
 MAD-QA     \citep{nguyen2024encoding}        & Movie videos         & Manual       & VideoQA & 19K              \\
 ActivityNet   \citep{yu2019activitynet}     &  YouTube videos     & Manual       & AL, TVR, VC, VMR & 5.8K              \\
 FIBER      \citep{castro2022fiber}        & VaTeX \citep{wang2019vatex}         & Manual       & VC, VideoQA      & 28K               \\
 WildQA    \citep{castro2022wild}         & YouTube videos         & Manual       & VideoQA          & 0.4K              \\
 NExT-QA  \citep{xiao2021next}          & VidOR \cite{shang2019annotating} & Manual       & VideoQA          & 5.4K              \\
 CausalVid-QA  \citep{li2022representation}     & Kinetics-700 \citep{carreira2019short} & Manual       & VideoQA          & 26K               \\
 HowTo100M  \citep{miech2019howto100m}        & YouTube videos         & Auto         & PT               & 1.2M              \\
 HD-VILA-100M \citep{xue2022advancing}      & YouTube videos         & Auto         & PT               & 3.3M              \\
 YT-Temporal-180M \citep{zellers2021merlot}  & YouTube videos         & Auto         & PT               & 6M                \\
 TGIF-QA  \citep{jang2017tgif}          & Animated GIFs & Manual       & VideoQA          & 71K               \\
 TGIF-QA-R   \citep{peng2021progressive}       & TGIF-QA \citep{jang2017tgif} & Manual, Auto & VideoQA          & 71K               \\
 DiDeMo    \citep{anne2017localizing}         & YFCC100M \citep{thomee2016yfcc100m} & Manual       & TVR              & 11K               \\
 YouCook2   \citep{zhou2018towards}        & YouTube videos         & Manual       & TVR, VC          & 2K                \\
HMDB-51  \citep{hmdb51}          & Web videos         & Manual       & TVR, AR          & 6.8K              \\
Kinetics-400 \citep{kinetics400}      & YouTube videos         & Manual       & AR               & 306K              \\
Kinetics-600  \citep{carreira2018short}     & Kinetics-400 \citep{kinetics400}         & Manual       & AR, VG           & 480K              \\
Kinetics-700  \citep{carreira2019short}     & Kinetics-600 \citep{carreira2018short}         & Manual       & AR               & 650K              \\
VaTeX    \citep{wang2019vatex}          & Kinetics-600 \citep{carreira2018short} & Manual       & TVR, VC          & 41K               \\
TVR     \citep{lei2020tvr}           & TVQA \citep{lei2018tvqa} & Manual       & VMR              & 22K               \\
How2R  \citep{li2020hero}            & HowTo100M \citep{miech2019howto100m} & Manual       & VMR              & 22K               \\
How2QA  \citep{li2020hero}           & HowTo100M \citep{miech2019howto100m} & Manual       & VideoQA          & 22K               \\
YouTube Highlights \citep{sun2014ranking} & YouTube videos         & Manual       & VMR              & 0.6K              \\
TACoS    \citep{regneri2013grounding}          & MPII Composites \citep{rohrbach2012database} & Manual       & VMR              & 0.1K              \\
QVHighlights  \citep{lei2021detecting}     & YouTube vlogs         & Manual       & VMR              & 10K               \\
TVSum    \citep{song2015tvsum}          & YouTube videos         & Manual       & VMR              & 50                \\
ViTT     \citep{huang2020multimodal}          & YouTube-8M \citep{youtube8m} & Manual       & VMR              & 5.8K              \\
VidChapters-7M  \citep{yang2023vidchapters}   & YT-Temporal-180M \citep{zellers2021merlot}         & Auto         & VC, VMR          & 817K              \\
VideoCC3M   \citep{nagrani2022learning}       & Web videos         & Auto         & PT               & 6.3M              \\
WebVid-10M  \citep{bain2021frozen}       & Web videos         & Auto         & PT               & 10.7M             \\
COIN      \citep{tang2019coin}         & YouTube videos         & Manual       & AS               & 12K               \\
CrossTask   \citep{zhukov2019cross}       & YouTube videos         & Manual       & AR               & 4.7K              \\
Alivol-10M   \citep{lei2021understanding}      & E-commerce videos         & Auto         & PT               & 10M               \\
LSMDC    \citep{rohrbach2015dataset}          & British movies         & Manual       & TVR              & 72                \\
EK-100   \citep{damen2022rescaling}          & Manual           & Manual       & AR, AL           & 7K                \\
SSV1     \citep{sthsth}          & Manual           & Manual       & AR               & 108K              \\
SSV2    \citep{sthsth}           & Manual           & Manual       & AR               & 221K              \\
Moments in Time  \citep{monfort2019moments}  & Web videos         & Manual       & AR               & 1M                \\
InternVid   \citep{wang2023internvid}       & YouTube videos         & Auto         & PT               & 7.1M              \\
How2      \citep{sanabria2018how2}         & YouTube videos         & Auto         & VC               & 13.2K             \\
WTS70M    \citep{stroud2020learning}         & YouTube videos         & Auto         & PT               & 70M               \\
Charades     \citep{gao2017tall}      & Manual           & Manual       & AR, VMR, VideoQA & 10K              \\ \hline
\end{tabular}}
\caption{Video understanding datasets in the literature. (VMR: Video moment retrieval, TVR: text-video retrieval, VC: video captioning, AL: action localization, AR: action recognition, AS: action segmentation, VG: video generation, PT: pre-training).}
\label{tab:datasets}
\end{table*}